\documentclass[sn-mathphys,Numbered]{sn-jnl}

\usepackage{graphicx}%
\usepackage{multirow}%
\usepackage{amsmath,amssymb,amsfonts}%
\usepackage{amsthm}%
\usepackage{mathrsfs}%
\usepackage[title]{appendix}%
\usepackage{xcolor}%
\usepackage{textcomp}%
\usepackage{manyfoot}%
\usepackage{booktabs}%
\usepackage{algorithm}%
\usepackage{algorithmicx}%
\usepackage{algpseudocode}%
\usepackage{listings}%

\theoremstyle{thmstyleone}%
\theoremstyle{thmstyletwo}%

\theoremstyle{thmstylethree}%

\usepackage[T1]{fontenc}

\begin{document}

\title[Article Title]{Transformer-based Named Entity Recognition in Construction Supply Chain Risk Management in Australia}


\author[1]{\fnm{Milad} \sur{Baghalzadeh Shishehgarkhaneh}}\email{milad.baghalzadehshishehgarkhaneh@monash.edu}

\author*[1,2]{\fnm{Robert C.} \sur{Moehler}}\email{robert.moehler@monash.edu}

\author[1]{\fnm{Yihai} \sur{Fang}}\email{yihai.fang@monash.edu}
\author[3]{\fnm{Amer A.} \sur{Hijazi}}\email{a.hijazi@ammanu.edu.jo}
\author[4]{\fnm{Hamed} \sur{Aboutorab}}\email{h.aboutorab@unsw.edu.au}

\affil[1]{\orgdiv{Civil Engineering}, \orgname{Monash University}, \orgaddress{\street{Clayton}, \city{Melbourne}, \postcode{3800}, \state{VIC}, \country{Australia}}}

\affil[2]{\orgdiv{Infrastructure Engineering}, \orgname{The University of Melbourne}, \orgaddress{\street{Parkville}, \city{Melbourne}, \postcode{3010}, \state{VIC}, \country{Australia}}}

\affil[3]{\orgdiv{Civil Engineering}, \orgname{Al-Ahliyya Amman University}, \orgaddress{\street{Al-Salt}, \city{Amman}, \postcode{19328}, \country{Jordan}}}
\affil[4]{\orgdiv{School of Business}, \orgname{UNSW Canberra}, \orgaddress{\street{Northcott Dr}, \city{Canberra }, \postcode{2600 }, \country{Australia}}}


\abstract{The construction industry in Australia is characterized by its intricate supply chains and vulnerability to myriad risks. As such, effective supply chain risk management (SCRM) becomes imperative. This paper employs different transformer models, and train for Named Entity Recognition (NER) in the context of Australian construction SCRM. Utilizing NER, transformer models identify and classify specific risk-associated entities in news articles, offering a detailed insight into supply chain vulnerabilities. By analysing news articles through different transformer models, we can extract relevant entities and insights related to specific risk taxonomies local (milieu) to the Australian construction landscape. This research emphasises the potential of NLP-driven solutions, like transformer models, in revolutionising SCRM for construction in geo-media specific contexts.}




\maketitle

\section{Introduction}\label{sec1} 
Natural Language Processing (NLP) is a field of computational techniques that aims to automate the analysis and representation of human language. By leveraging both theoretical principles and practical applications, NLP enables us to work with natural language data in various ways. From parsing and part-of-speech (SOP) tagging to machine translation, conversation systems, and named entity recognition (NER), NLP encompasses a wide range of components and levels. It has proven itself useful in fields such as natural language understanding, generation, voice/speech recognition, spell correction, grammar check, among others. The versatility of NLP allows it to address diverse linguistic tasks effectively \cite{bib1}.

The evolution of NLP can be divided into different phases that represent the progress in language generation and other language processing aspects. These phases illustrate the current state of the field, as well as ongoing trends and challenges. NLP encompasses a wide range of applications and continues to advance with computational modelling and technological innovations \cite{bib2}. Furthermore, NLP involves studying mathematical and computational models related to various language aspects. It includes developing systems like spoken language interfaces that combine speech and natural language, as well as interactive interfaces for databases and knowledge bases. This enables modelling of human-human and human-machine interactions. Overall, NLP is a multidisciplinary field that intertwines computational, linguistic, and cognitive dimensions \cite{bib3,4}.

A dedicated series focused on the "Theory and Applications of Natural Language Processing" explores the latest advancements in computational modelling and processing of speech and text in different languages and domains \cite{bib5}. This highlights the rapid progress in NLP and Language Technology, driven by the increasing volume of natural language data and the evolving capabilities of machine learning and deep learning technologies. These references illustrate that NLP is a dynamic field with a solid theoretical foundation, powering numerous practical applications across various domains. NER is a method for identifying, classifying, and separating named entities into groups according to predetermined categories. NER is a crucial component of NLP technology and forms the foundation for many studies in this field. The recent advancements in deep learning have significantly improved the performance of NER applications, especially in real-world situations where high-quality annotated training data is often limited \cite{bib6}.

In the financial sector, NER plays a crucial role in extracting important information from unstructured data. This process is essential for various analytical and decision-making processes within finance \cite{bib7,8}. Furthermore, NER in the medical field has multiple applications. It aids in clinical decision support, analysing medical literature, managing electronic health records (EHR) \cite{bib9}, detecting relationships between entities, extracting valuable information from text data \cite{bib10}, mining and analysing text documents for useful information, and facilitating drug discovery, treatment planning, and disease monitoring by identifying and categorizing medical entities like drugs, treatments, and diseases \cite{bib11}. Considering construction industry, construction management has benefited from the global use of NER in automating and improving various processes. For example, NER models have been used to automatically extract information from construction specifications, particularly in road construction projects, aiding in bid management \cite{bib12}. In Chinese construction documents, NER helps identify common tasks and ensures consistent annotation, which is crucial for improving efficiency \cite{bib13}. Additionally, NER has been employed to identify building construction defect information from residents' complaints, demonstrating its potential in defect management \cite{bib14}. These applications highlight the versatility of NER in addressing diverse challenges within construction management and providing a solid foundation for enhancing efficiency in this field. Advancements in machine learning (ML) and NLP technologies have greatly influenced the development of NER methodologies. In the early stages, NER tasks mainly relied on rule-based and dictionary-based methods \cite{bib15,16}. These approaches involved using manually created rules and predefined dictionaries to detect and categorize named entities in text. As ML emerged, researchers started using ML models like Hidden Markov Models (HMM), Decision Trees, Maximum Entropy Models (ME), and Support Vector Machines (SVM) to enhance the performance of NER. These statistical methods proved to be effective in improving the accuracy of NER tasks \cite{bib17}. As computing and algorithms advanced, Conditional Random Fields (CRF) became the preferred choice for NER. CRFs are particularly suitable for sequence labelling tasks because they can consider the correlation between neighboring sequences, unlike generative models like HMM \cite{bib18}. Maximum Entropy models also started being used around this time, although they had the issue of label bias. However, CRFs were able to address this problem by jointly considering the weights of different features across all states, rather than normalizing transition probabilities at the state level \cite{bib19}. 

\begin{figure}[t!]
    \centering
    \includegraphics[width=1\textwidth]{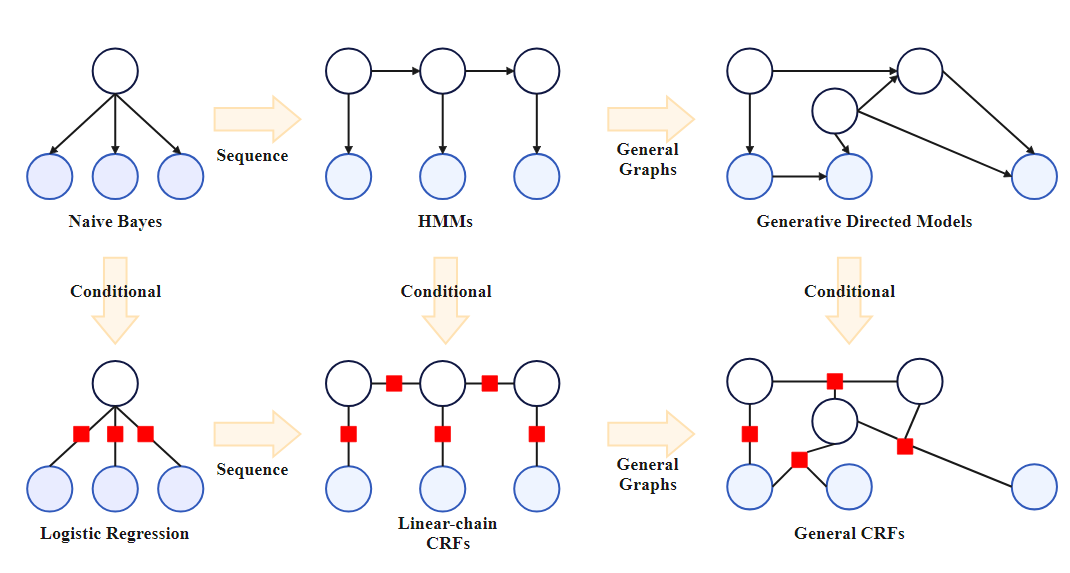}
    \caption{\textbf{Evolution of Probabilistic Models in Machine Learning.}}
   \label{tab:myfig1}
\end{figure}

However, there has been a shift in the paradigm with the emergence of deep learning techniques driven by neural networks. These methods have shown great success in tasks such as NER, as confirmed by several studies \cite{bib20}. One particularly acclaimed approach combines Long Short-Term Memory (LSTM) with CRF. In this combination, LSTM carefully captures vector representations of each word or token in a sentence, which are then fed into the CRF model for accurate sequence tagging \cite{bib21}. In a ground-breaking approach, \citeauthor{bib22} (\citeyear{bib22}) combined character-level and word-level features in a hybrid network architecture. Their model utilized a BiLSTM layer followed by a log-SoftMax layer to independently decode each tag, resulting in improved accuracy. Similarly, \citeauthor{bib23} (\citeyear{bib23})  merged CRFs with information entropy, effectively identifying abbreviations of financial named entity candidates. This demonstrates the versatility of such models in specialized domains. \hyperref[tab:myfig1]{Fig. 1} shows the evolution of probabilistic models in machine learning.

To contribute to the ongoing discussion, \citeauthor{bib24} (\citeyear{bib24})  introduced a novel approach that combined a BiLSTM encoder with an incrementally decoded neural network structure. This innovative method allowed for simultaneous decoding of tags, promoting a more nuanced comprehension of textual data. Although there were various encoding strategies based on recurrent neural network (RNN) architectures, the differences in methodology became evident during the decoding phase. Recently, advanced language models like ELMo \cite{bib25}, GPT-4 \cite{bib26}, and BERT \cite{bib27}, have emerged in the field of technology. These models have become extremely effective across various NLP tasks and have revolutionized the way we approach natural language processing. Unlike traditional methodologies that heavily relied on feature engineering, these deep neural networks possess the remarkable ability to automatically extract features from data. This characteristic has propelled them to achieve superior performance without the need for manual feature crafting or extensive domain expertise. The adoption of these sophisticated models marks a significant milestone in addressing NER tasks, facilitating more efficient and automated approaches in identifying and categorizing named entities across diverse domains and languages.

\section{Literature Review}\label{sec2}
The use of advanced language models, like BERT and GPT-3, in NER has become increasingly prevalent across various industries. From healthcare to finance, legal, and construction, businesses are leveraging these sophisticated models to accurately identify and categorize named entities within large volumes of text. These models have the remarkable ability to autonomously detect complex patterns and relationships between words without the need for labour-intensive feature engineering. This capability allows for a nuanced understanding of data, enabling critical insights extraction, better decision-making, regulatory compliance, and improved customer experiences. Additionally, advancements in transfer learning and the development of domain-specific pre-trained models have further accelerated the effectiveness and adoption of NER across diverse industries. In today's data-driven ecosystem, NER has become an indispensable tool.

\citeauthor{bib28} (\citeyear{bib28}) utilized Chinese Electronic Health Record (EHR) datasets to evaluate different models for NER. Their findings revealed that the BERT-BiLSTM-CRF model outperformed other models such as BiLSTM and word2vec. With an impressive F1 score of approximately 75\%, this model proved highly effective in extracting medical information from extensive EHRs. \citeauthor{bib29} (\citeyear{bib29}) created a NER methodology to identify Chinese medicine and disease names in conversations between humans and machines. They evaluated various models, and the combination of RoBERTa with biLSTM and CRF performed the best. Using a corpus obtained through web crawling, this model achieved an impressive Precision, Recall, and F1-score of 0.96. These findings highlight its potential for enhancing medication reminders in dialogue systems. \citeauthor{bib30} (\citeyear{bib30}) developed a Chinese NER model called BBIEC specifically for analysing COVID-19 epidemiological data. This model effectively processes unlabelled data at the character level, extracting global and local features using pre-trained BERT, BiLSTM, and IDCNN techniques. The BBIEC model outperforms traditional models when it comes to recognizing entities that are crucial for analysing the transmission routes and sources of the epidemic. \citeauthor{bib31} (\citeyear{bib31}) proposed a BERT-Transformer-CRF based service recommendation method (BTC-SR) for enhanced chronic disease management, which initially employs a BERT-Transformer-CRF model to identify named entities in disease text data, extracts entity relationships, and integrates user implicit representation to deliver personalized service recommendations, demonstrating improved entity recognition with an F1 score of 60.15 on the CMeEE dataset and paving the way for more precise service recommendations for chronic disease patients.

\citeauthor{bib32} (\citeyear{bib32}) introduced a deep learning-based Mineral Named Entity Recognition (MNER) model, utilizing BERT for mineral text word embeddings and enhancing sequence labelling accuracy by integrating the CRF algorithm's transfer matrix. Furthermore, \citeauthor{bib33} (\citeyear{bib33}) introduced a multi-task model called BERT-BiLSTM-AM-CRF. The model utilizes BERT for dynamic word vector extraction and then refines it through a BiLSTM module. After incorporating an attention mechanism network, the output is passed into a CRF layer for decoding. The authors tested the model on two Chinese datasets and observed significant improvements in F1 score compared to previous single-task models, with increases of 0.55\% in MASR dataset and 3.41\% in People's Daily dataset respectively. 
\citeauthor{bib34} (\citeyear{bib34}) explored the NER task in Telugu language using various embeddings such as Word2Vec, Glove, FastText, Contextual String embedding, and BERT. Remarkably, when combining BERT embeddings with handcrafted features, the results outperformed other models significantly. The achieved F1-Score was an impressive 96.32\%. \citeauthor{bib35} (\citeyear{bib35}) introduced Wojood, a unique corpus specifically designed for Arabic nested NER. This corpus comprises approximately 550K tokens of Modern Standard Arabic and dialect, each manually annotated with 21 different entity types. Unlike traditional flat annotations, Wojood includes around 75K nested entities, accounting for about 22.5\% of the total annotations. The accuracy and reliability of this corpus are evident in its substantial interannotator agreement, with a Cohen's Kappa score of 0.979 and an F1 score of 0.976. Furthermore, to address the limitations of traditional methods for named entity recognition in the context of agricultural pest information extraction, \citeauthor{bib36} (\citeyear{bib36}) proposed a PBERT-BiLSTM-CRF model. This model leverages pre-trained BERT to resolve ambiguity, BiLSTM to capture long-distance dependencies, and CRF for optimal sequence annotation. The results demonstrate significant improvements in precision, recall, and an impressive F1 score of 90.24\% compared to other models. 

\subsection{Named Entity Recognition in Construction Industry}\label{subsec2}
Named entity recognition in construction has received some attention in academic literature, although the available published research in this field is relatively limited. While several studies have been conducted on this topic, the quantity of publications compared to other areas of natural language processing and construction is modest. In the realm of Construction Supply Chain Risk Management (CSCRM) in Australia, the significance of local and international news cannot be overstated. 

The constantly changing geopolitical, environmental, and economic scenarios greatly impact construction supply chains. For example, the recent disruptions caused by the COVID-19 pandemic had a profound effect on the China-Australia construction supply chain. This highlighted the urgent need for timely and accurate information to effectively manage and mitigate risks \cite{bib37}. The construction sector in Australia is currently facing increased supply chain risks. These risks have been amplified by the growing number of suppliers, complex work streams, stringent compliance requirements, and difficulties in finding eligible parties. It is important to note that disruptions in global supply chains, particularly those originating from regions like China, have resulted in project delays. This emphasizes the significance of international news for predicting and managing such disruptions. The lack of transparency in supply chain risk among Australian construction firms emphasizes the need for a well-informed and data-driven approach to risk management. By utilizing NER technologies, particularly in the context of geological news texts, automation can play a vital role in extracting relevant information from local and international news sources. This enhancement significantly improves the accuracy and time­liness of risk assessments and mitigating actions within the Australian construction supply chain domain. However, the field of geological news texts is rapidly expanding, offering a wealth of valuable information. Accurately extracting this information can greatly enhance geological survey efforts. However, traditional manual extraction methods are inefficient and time-consuming, leading to lower accuracy. As the volume of geological news text data increases, these challenges become even more pronounced. It is crucial to transition towards automated extraction paradigms to address this complexity. Automating the extraction of geological news entities goes beyond just a procedural evolution; it represents a fundamental leap towards the creation of comprehensive geological knowledge graphs. These knowledge graphs can serve as structured repositories, facilitating the retrieval and analysis of geological information and propelling advancements in the field of geological surveys.

The recent advancements in machine learning and NLP have significantly improved the challenges associated with manual data extraction. One notable breakthrough is the emergence of transformer-based models like BERT, which has paved the way for automating the extraction process. For example, a study introduced a method called Geological News Named Entity Recognition (GNNER) that utilizes the BERT language model to effectively extract and leverage geological data \cite{bib38}. Moreover, other scholarly endeavours have demonstrated automated techniques for extracting spatiotemporal and semantic information from geological documents. These techniques are crucial for tasks such as data mining, knowledge discovery, and constructing knowledge graphs \cite{bib39,40}. The narrative above explains the importance and modern approaches used in automating the extraction of geological news information. This automation not only enhances the efficiency and accuracy of retrieving information, but it also forms a vital foundation for building comprehensive geological knowledge graphs.

The integration of NER technologies, such as advanced models like BERT, into CSCRM frameworks can greatly enhance the automatic extraction of critical information entities from a vast amount of news data. This, in turn, enables the creation of comprehensive knowledge graphs that encompass various risk factors and their potential impact on construction supply chains. These knowledge graphs hold immense value for construction firms, regulators, and other stakeholders as they foster a more resilient, transparent, and responsive ecosystem within the Australian construction supply chain. Despite its wide-ranging applications, there seems to be a dearth of research or documentation on the use of NER in construction supply chain risk management, particularly with regards to geological news in Australia. This presents an opportunity for further investigation and exploration into utilizing NER to address risk management challenges within the construction supply chain domain. Specifically, it can prove valuable in leveraging geological news for more informed decision-making processes.
This research study examined the effectiveness of various BERT models in performing NER within the field of Construction Supply Chain Risk Management (CSCRM). The primary source of information utilized is news data. This investigation breaks new ground by exploring NER applications in CSCRM specifically through the lens of news data, an area that hasn't been previously studied. The dataset consists of information gathered from multiple news outlets, providing a fertile ground for identifying and analysing numerous risk factors and how they manifest within the construction supply chain ecosystem. Through careful examination, this study uncovers several significant contributions. Firstly, it establishes a practical framework for utilizing NER to dissect real-world news data and extract valuable risk-related entities and their relationships. This contributes to a deeper understanding of risk dynamics in construction supply chains. Secondly, this research provides a comparative analysis of different BERT models in accurately discerning these entities. This serves as a solid foundation for further advancements in the field. Lastly, the insights obtained through our analysis pave the way for developing more resilient and informed risk management strategies in the construction sector. It represents a significant step towards mitigating vulnerabilities within supply chains through the effective utilization of NER technologies.

\section{Materials and Methods}\label{sec3}

\subsection{WORD2VEC}\label{subsec2}
The field of semantic vector spaces has evolved through the use of neural models, building upon previous foundational work \cite{bib41}. While there are many word embedding models available, one prevailing paradigm that utilizes neural networks is known as Word2Vec (W2V) \cite{bib42, 43}. The W2V model generates word embeddings using two main methodologies: Continuous Bag-of-Words (CBOW) and Skip-Gram (SG). These techniques have different ways of managing input and output variables but share a similar network structure. Researchers have also expanded on the original W2V model to create multi-sense word embeddings. This aims to improve how different senses of a word are represented through clustering mechanisms. Additionally, there is a growing body of research focusing on contextualized word embedding algorithms, including some well-known models in this area like Elmo, Bert, and Xlnet. These models reflect the broader trend in NLP towards achieving enhanced semantic understanding \cite{bib44}.

Word embeddings are a hot topic in contemporary discussions, particularly with regard to the influential W2V model. Within the W2V library, both the SG and CBOW models have played instrumental roles in shaping word embedding methodologies. These models resemble shallow neural networks, similar to the original Simple Recurrent Network (SRN), and have gained recognition for their ability to predict neural network models. They bridge the gap between count-based distributional semantics models without any notable disparity in quality \cite{bib45}. The introduction of W2V has not only propelled the field towards a more nuanced exploration of semantic spaces but also provided a strong foundation for the wide­spread use of pre-trained language models. As a result, it has significantly amplified the effectiveness and range of text analytic tasks within the realm of deep learning \cite{bib46}. 

\subsection{TRANSFORMERS}\label{subsec2}
\begin{figure}[t!]
    \centering
    \includegraphics[width=1\textwidth]{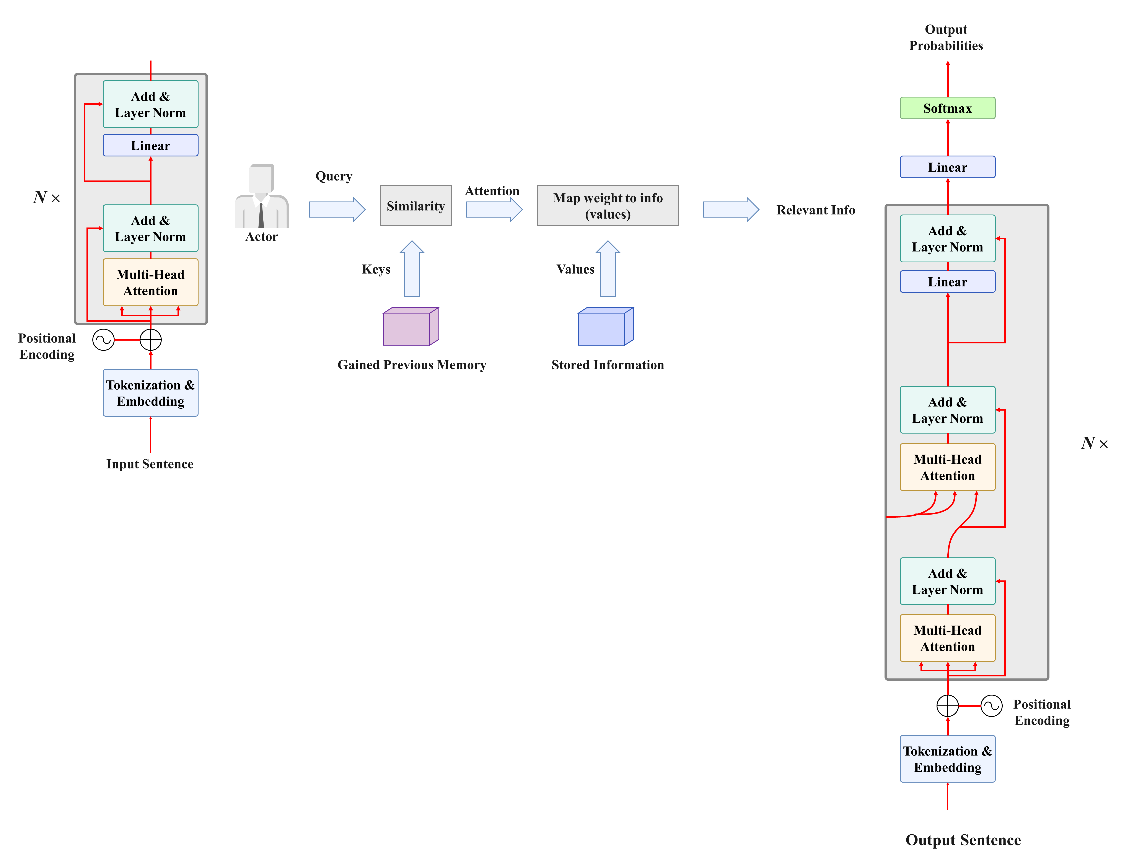}
    \caption{\textbf{Transformer’s Architecture.}}
   
    \label{tab:myfig}
\end{figure}

The Transformer architecture, as described in the influential work by \citeauthor{bib47} (\citeyear{bib47}), presents a unique framework for transferring weighted knowledge between different neural components. Unlike traditional approaches that rely on recurrent or convolutional structures, the Transformer exclusively utilizes attention mechanisms. At the heart of its effectiveness is the attention mechanism, which assigns weights to each input representation and learns to focus on important segments of the data. The output is then computed by taking a weighted sum of these values, with the weights determined through evaluating how well the query matches with its corresponding key \cite{bib47}. The innovative design of the Transformer has not only advanced NLP but also made significant progress in computer vision and spatio-temporal modelling \cite{bib48, 49}. By efficiently processing sequential data like sentences, the Transformer has improved model performance through enhanced parallelization, reducing training times \cite{bib50}. Its attention-centric mechanism enables the architecture to capture global dependencies between input and output, pushing the boundaries of what can be achieved in NLP and related fields. \hyperref[tab:myfig]{Fig. 2} represents the Transformers architecture. 

BERT (Bidirectional Encoder Representations from Transformers), however, is a major breakthrough in the field of deep language understanding. Its architecture, which utilizes the powerful Transformer model, particularly its encoder component, has revolutionized our ability to comprehend natural language. BERT's pre-training phase involves analysing an enormous corpus of books and Wikipedia articles, allowing it to grasp the complex semantics present in textual data. The core essence of BERT lies within the encoder section of the Transformer model—an innovative design introduced by \citeauthor{bib47} (\citeyear{bib47}),—which has received widespread acclaim for its efficient parallelization of computations, greatly improving computational efficiency. The BERT model takes in input as a sequence of tokens, with special tokens [CLS] and [SEP] indicating the beginning and end of sequences. Within this token sequence, there are three types of embeddings that play a key role in helping the model understand the text: token embeddings, segment embeddings, and position embeddings. These embeddings are crucial for the model's language comprehension abilities.

\textbf{Token Embedding:} Token embeddings are crucial for identifying and representing words or sub-words in the input. Each token in the input sequence is assigned a specific embedding, which represents it in a high-dimensional space. In BERT, every token in the WordPiece token vocabulary has its own learned token embeddings during training \cite{bib51}.

\textbf{
Segment Embedding:} BERT uses segment embeddings to distinguish between different sentences, particularly when it processes pairs of sentences for tasks like question answering. It learns separate embeddings for the first and second sentences, allowing the model to differentiate between them effectively \cite{bib51}.

\textbf{Position Embedding:} In models like BERT, which do not have a recurrent structure, position embeddings play a crucial role in understanding the order of words in a sentence. Unlike traditional recurrent networks such as LSTMs, BERT relies on position embeddings to provide the necessary positional information. Different methods have been proposed to model word order in Transformer-based architectures like BERT \cite{bib52}.
Figure 3 provides a clear visual representation of the Transformer model's architecture, with a specific focus on the encoder segment. This segment plays a significant role in the widely used BERT model. The composition and operational principles of this crucial encoder segment are as follows:
\subsubsection{TRANSFORMER’S INPUT}\label{subsubsec2}

The Transformer model processes all tokens in a sequence simultaneously. In order to preserve the positional information of each token, a positional encoding is added to every word vector, as shown in (1) as follows:
\begin{equation}X=embLookup(X)+PosEncoding .\label{eq}\end{equation}

where embLookup(X) is used to retrieve the embedding of each token X in the sequence. Additionally, PosEncoding is performed to add a unique positional encoding to each token's embedding. This positional encoding is essential because it enables the model to understand the order of tokens, which is crucial for tasks like NER \cite{bib53}.

\subsubsection{SELF-ATTENTION MECHANISM}\label{subsubsec2}

Unlike the traditional attention mechanism, the self-attention mechanism focuses on capturing relationships within the input or output sequence. This makes it an advanced version of the attention mechanism. The self-attention process involves three matrices: Query (Q), Key (K), and Value (V). In information retrieval systems, Q represents the input information, K represents relevant content that matches Q, and V refers to the actual information itself. In the Transformer model, each layer consists of a self-attention module followed by a feed-forward layer. Additionally, each layer includes layer normalization and residual connections as additional components. The self-attention mechanism in the Transformer model allows for the computation of attention scores. These scores determine the level of focus each token should have on other tokens in the sequence. This mechanism is essential for handling sequences and has played a key role in the success of the Transformer model in various NLP applications \cite{bib54}. BERT’s architecture is based on a transformer model, specifically an "encoder-only" version. It consists of two main components: an embedding module and a series of encoders, which are essentially the same as Transformer encoders. These encoders enable BERT to efficiently process and analyse input text, resulting in its impressive performance across various NLP tasks.

To use this mechanism, we start with matrix X, and then derive Q, K, and V from X using linear transformations (as shown in (2-4)). Matrix x is the word vector matrix that interacts with auxiliary matrices WQ, WK, and WV to obtain the corresponding Q, K, and V values for each item in the sequence. Next, the current item's Q is compared to each item's K in the sequence to determine their relationship. After scaling and normalizing the product using Softmax, it is multiplied by V. The resulting V values are then aggregated to determine the feature representation of the current item as described in (5).
\begin{equation}
Q = \text{Lin}(X) = XW_Q
\label{eq2}
\end{equation}
\begin{equation}
Q = \text{Lin}(X) = XW_K
\label{eq2}
\end{equation}
\begin{equation}
Q = \text{Lin}(X) = XW_V
\label{eq2}
\end{equation}
\begin{equation}
X_{\text{att}} = \text{SelfAtt}(Q, K, V) = \text{Softmax}\left(\frac{QK^T}{\sqrt{d_k}}\right)V
\tag{5}
\label{eq5}
\end{equation}
where \( d_k \) shows the dimensional of the Q and K vectors.
\subsubsection{MULTI-HEAD MECHANISM }
In the self-attention mechanism, each item in the input sequence is assigned three feature expressions: Query, Key, and Value. On the other hand, the multi-head mechanism creates multiple sets of auxiliary matrices within the Transformer model and combines them with the word vector input matrix X to obtain several sets of Query, Key, and Value values. This means that every item in the sequence is connected to multiple sets of feature expressions. These sets are then concatenated and passed through a fully connected layer for dimensional reduction. This mechanism allows the model to effectively recognize different positional and contextual relationships among tokens simultaneously, thereby enhancing its ability to represent information accurately.

\subsubsection{NORMALISATION AND SUMMATION}
To improve feature extraction, it is necessary to include a residual connection. This connection combines the vector from the self-attention and multi-head mechanisms with the original input vector, as shown by (6-7).  This ensures that important information is preserved throughout the process. Additionally, implementing hidden layer normalisation plays a key role in speeding up convergence, allowing for more efficient training of the model.
\begin{equation}
X_{\text{att}} = X + X_{\text{att}}
\tag{6}
\label{eq6}
\end{equation}

\begin{equation}
X_{\text{att}} = \text{LayerNorm}(X_{\text{att}})
\tag{7}
\label{eq7}
\end{equation}
However, in addition to the BERT model, this paper also investigates other models like RoBERTa, DistilBERT, ALBERT, ELECTRA, T5, and GPT-3. Each of these models has specific adjustments and enhancements designed for different NLP tasks.

RoBERTa, also known as Robustly Optimized BERT Pre-training Approach, enhances the performance of BERT by modifying the pre-training process. This includes longer training periods, the use of larger datasets, and bigger mini-batches compared to BERT \cite{bib55}. Furthermore, DistilBERT is a more compact and efficient version of BERT. It was created using a process called knowledge distillation, where the DistilBERT model learns from a pre-trained BERT model. This allows DistilBERT to maintain similar performance capabilities while being faster and more economical in terms of computational resources \cite{bib56}. To make BERT more efficient, ALBERT (A Lite BERT) utilizes techniques like factorised embedding parameterisation and cross-layer parameter sharing. These methods reduce the size and increase the speed of BERT \cite{bib57}. Instead of using the masked language modelling objective like BERT, ELECTRA (Efficiently Learning an Encoder that Classifies Token Replacements Accurately) takes a different approach. It utilizes a pre-training task called replaced token detection, which aims to achieve more efficient pre-training \cite{bib58}. T5 (Text-To-Text Transfer Transformer) takes a distinctive approach by transforming every NLP problem into a text-to-text format. This simplifies the application of the model to various NLP tasks \cite{bib59}. Finally, the GPT (Generative Pretrained Transformer) model is a ground-breaking technology that has revolutionised NLP. Through unsupervised learning on vast amounts of text data, it has successfully generated text that closely resembles human writing \cite{bib60}. GPT-3 is the successor to GPT-2 and boasts a significant raise in both parameter count (from 1.5 billion to 175 billion) and data processed (from 40 GB to 45 TB), making it the largest language model ever created \cite{bib61}.

\subsection{DATA GATHERING}\label{subsec2}
This study conducted a thorough investigation to create a detailed risk categorisation specifically for managing risks in the construction supply chain in Australia. This involved carefully reviewing existing literature and incorporating insights from the Cambridge Taxonomy of Business Risks \cite{bib62}. The resulting risk categorisation covers a wide range of risks commonly found in the Australian construction supply chain, providing a strong basis for the following stages of this study.

After establishing the risk taxonomy, the attention turned to collecting a comprehensive dataset for thorough analysis of the identified risks. A specialised News API was utilised to search through approximately 2000 articles from renowned news sources like The Australian, Sky News Australia, Bloomberg, CNN, Reuters, and Google News. This data collection approach was carefully designed to adhere to web scraping guidelines and ensure ethical acquisition of data. The result was a diverse and extensive dataset that provided ample material for empirical investigation of the specified risks within the CSCRM domain using NER.
\subsubsection{ANNOTATION OF TEXT CORPUS }
For dataset annotation, sequence labelling is a critical step that helps organise data for further analysis. Among various labelling methods used in scholarly research, this study utilizes the "BIO" labelling scheme due to its effectiveness and widespread acceptance. This labelling convention, commonly employed in NER, offers a systematic approach to annotate text sequences, allowing for a detailed understanding of the text structure. The "BIO" labelling scheme consists of three annotations: "B-X," "I-X," and "O." In this scheme, the letter "B" indicates the beginning of a named entity in the text. The letter "I" represents the middle and concluding segments of the named entity. Lastly, the letter "O" denotes text segments that do not contain a named entity \cite{bib63}.

After carefully applying the "BIO" labelling scheme to the news texts, we were able to obtain a substantial data-set with labelled information. Our statistical analysis after labelling revealed an impressive count of 39,500 entities across six different categories, as shown in \hyperref[tab:mytable]{Table 1}. Figure 4 shows the methodology of the current research work.
\begin{table}[h]
\caption{\textbf{Annotated Corpus's Entities Number}}\label{tab1}
\setlength{\tabcolsep}{3pt}
\begin{tabular}{@{}lll@{}}
\toprule
Entity & Category & Occurrences \\
\midrule
PER & Person’s names & 560 \\
RRE & The most relevant risk events from risk taxonomy & 3674 \\
PNR & Political, Nationalities, and Religious groups & 570 \\
OSC & Organisations, Suppliers, and Companies & 1416 \\
GPU & Geo Political Units & 3606 \\
CMS & Construction Materials & 570 \\
\botrule
\label{tab:mytable}
\end{tabular}
\end{table}

\begin{figure}[t!]
    \centering
    \includegraphics[width=1\textwidth]{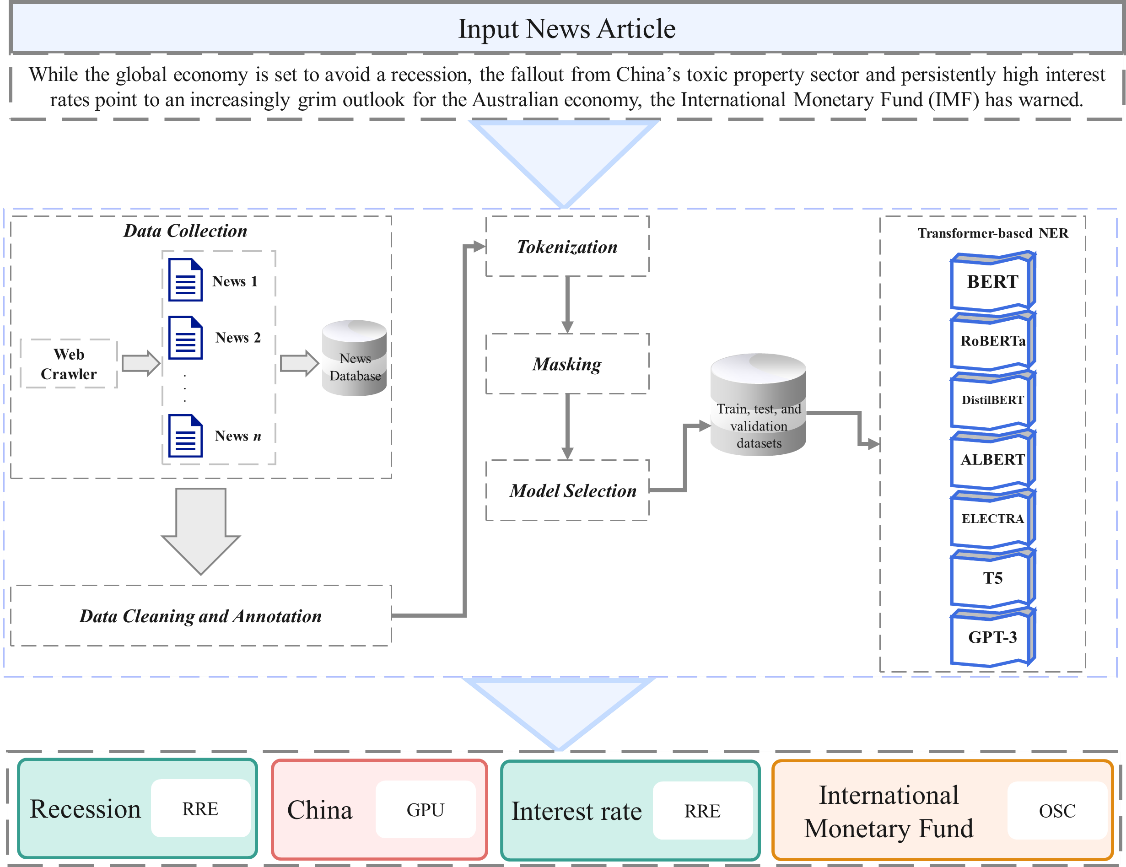}
    \caption{\textbf{Transformer’s Architecture.}}
    \label{fig1}
\end{figure}

In order to train the transformers models, it requires powerful computational resources. Table 2 provides detailed information about the hardware and software used in this experiment, giving a comprehensive understanding of the infrastructure that supported the training of the transformer models in this study.

\begin{table}[h]
\caption{\textbf{Experimental Setup of This Study}}\label{table2}
\setlength{\tabcolsep}{3pt}
\begin{tabular}{@{}lll@{}}
\toprule
\textbf{Type} & \textbf{Configuration} & \textbf{Features} \\
\midrule
\multirow{6}{*}{Software} & CUDA & 11.5 \\ 
 & Python & 3.9 \\ 
 & Numpy & 1.21.2 \\ 
 & Scikit-learn & 0.24.2 \\ 
 & Pandas & 1.3.3 \\ 
 & TensorFlow & 2.6 \\ 
 & PyTorch & 1.9.0 \\ 
\midrule
\multirow{4}{*}{Hardware} & Operating System & CentOS Version 8.4 \\ 
 & Video RAM & 11 Gigabytes GDDR6 \\ 
 & RAM & 32.0 Gigabytes \\ 
 & Processor & Intel Core i7-12700, 12th Generation \\ 
\botrule
\end{tabular}
\end{table}

\begin{table}[h]
\caption{\textbf{Models' Hyperparameters}}\label{table3}
\setlength{\tabcolsep}{3pt}
\begin{tabular}{@{}ll@{}}
\toprule
\textbf{Hyperparameters} & \textbf{Values} \\
\midrule
Drop Rate & 0.50 \\
lr & 3e-5 \\
Batch Size & 32 \\
Epochs & 10 \\
Max len & 128 \\
\botrule
\end{tabular}
\end{table}

During the model training phase, the choice of hyper-parameters greatly affects the outcomes. To ensure consistency and reduce variability between experiments, this study used a fixed set of hyper-parameters for training different models. The important parameters involved in the training process are listed in Table 3. In this table, an epoch refers to one complete iteration over the entire training dataset, max len indicates the maximum sequence length, batch size determines how much data is processed in each training iteration, lr controls the rate of learning, and drop rate helps prevent over-fitting in the neural network.

\section{Results and Discussion}\label{sec4}
This section discusses the results of different models that were used for NER in news articles focusing on construction supply chain risk management.

\subsection{EXPERIMENT DESIGN AND ASSESSMENT}\label{subsec2}
When evaluating the performance of different models in NER, precision (P), recall (R), and F1-score (F1) are commonly used metrics. These metrics help assess how well the models perform. The specific computational formulas for these metrics are as follows:
\begin{equation}
P = \frac{\text{TP}}{\text{TP} + \text{FP}}
\tag{8}
\label{eq8}
\end{equation}

\begin{equation}
R = \frac{\text{TP}}{\text{TP} + \text{FN}}
\tag{9}
\label{eq9}
\end{equation}

\begin{equation}
F1 = 2 \times \frac{P \cdot R}{P + R}
\tag{10}
\label{eq10}
\end{equation}

where \textit{TP} represents the number of correctly identified entities or true positives. \textit{FP} denotes the number of incorrectly identified entities or false positives. Lastly, \textit{FN} signifies the number of missed entities or false negatives.
\subsection{MODEL EVALUATION AND COMPARISON}
For our study, we carefully divided the labelled data-set into three separate subsets: the training set, validation set, and test set. We followed a distribution ratio of 8:1:1 to allocate the entities. This means that we had 31,600 entities for training, 3,950 entities for validation, and another 3,950 entities for testing. This division is crucial in order to train and evaluate models effectively and ensure their robustness and ability to generalise in line with suggestions from \citeauthor{bib12} (\citeyear{bib12}). To study NER in CSCRM, we trained seven different models using a designated training data-set. After training, we evaluated the performance and effectiveness of these models on a separate test set for NER tasks. This approach is similar to the method used by when evaluating multiple models to determine the best one for NER tasks in a similar domain \cite{bib14}. Precision, Recall, and F1 scores of he mentioned models are shown in Table 4.

 In the provided evaluation metrics (Precision, Recall, and F1 Score), Table 4 presents a comparative analysis of the models. RoBERTa stands out with an impressive average F1 score of 0.8580, which indicates a well-rounded performance in both precision (0.9341) and recall (0.8023). On the other hand, T5 exhibits the highest average precision value of 0.9924 but suffers from a low recall of 0.3645, resulting in a modest F1 score of 0.5115. These differences highlight varying capabilities among the models in accurately identifying entities and retrieving relevant instances from the news dataset.

When evaluating the performance of models in various categories such as PER, RRE, PNR, OSC, GPU, and CMS, it becomes evident that each model has its strengths and weaknesses. In terms of precision, almost all models demonstrate high accuracy in the PNR and CMS entities. Some even achieve a perfect score of 1.0000. However, the OSC entity poses challenges for all models. Though T5 exhibits the highest precision score of 1.0000 in this category, its recall rate is notably lower. This suggests that factors like entity characteristics or variations in training data quality and quantity significantly impact the overall performance of these models across different categories. The performance of models in NER tasks is significantly influenced by their underlying architectures and training data. Transformer-based models like BERT, RoBERTa, and DistilBERT excel in capturing contextual relationships, which are crucial for NER tasks. On the other hand, models like T5 and GPT-3 approach NER differently as text-to-text and generative models, respectively.

GPT-3's performance in NER tasks is generally lower than supervised baselines due to the inherent gap between NER (a sequence labelling task) and GPT-3's nature as a text generation model. However, adaptations such as GPT-NER have been proposed to bridge this gap by transforming the sequence labelling task into a generation task that can be easily tailored for large language models like GPT-3 \cite{bib60}.
Moreover, ELECTRA uses a unique pre-training task where token detection is replaced with distinguishing "real" from "fake" input data. This can potentially improve its NER performance by reducing false positives and negatives in entity recognition \cite{bib64}. When evaluating and selecting models for implementation within the construction supply chain domain, it is crucial to consider both the architectural differences of the models and the nature of the NER tasks. This analysis highlights the significance of this dual consideration.

\begin{table*}[htbp]
\caption{Statistical Results of Different Models for NER in CSCRM}
\label{table:ner_results}
\centering
\setlength{\tabcolsep}{3pt}
\begin{tabular}{|l|l|l|l|l|l|l|l|l|}
\hline
\textbf{Models} & \textbf{Evaluation} & \textbf{PER} & \textbf{RRE} & \textbf{PNR} & \textbf{OSC} & \textbf{GPU} & \textbf{CMS} & \textbf{Average} \\
\hline
BERT & P & 0.9565 & 0.9813 & 1.0000 & 0.7424 & 0.9663 & 0.9545 & 0.9335 \\
     & R & 0.5789 & 0.8739 & 0.6176 & 0.8033 & 0.9306 & 0.8235 & 0.7713 \\
     & F1 & 0.7213 & 0.9244 & 0.7636 & 0.7717 & 0.9484 & 0.8842 & 0.8356 \\
\hline
RoBERTa & P & 0.8667 & 0.9807 & 1.0000 & 0.8116 & 0.9808 & 0.9649 & 0.9341 \\
        & R & 0.6341 & 0.8841 & 0.6444 & 0.8358 & 0.9143 & 0.9016 & 0.8023 \\
        & F1 & 0.7324 & 0.9299 & 0.7838 & 0.8235 & 0.9464 & 0.9322 & 0.8580 \\
\hline
DistilBERT & P & 0.8696 & 0.9789 & 0.9474 & 0.7544 & 0.9746 & 1.0000 & 0.9208 \\
           & R & 0.5405 & 0.8722 & 0.5625 & 0.7679 & 0.9412 & 0.8696 & 0.7589 \\
           & F1 & 0.6667 & 0.9225 & 0.7059 & 0.7611 & 0.9576 & 0.9302 & 0.8240 \\
\hline
ALBERT & P & 0.9200 & 0.9763 & 0.7000 & 0.6935 & 0.9787 & 1.0000 & 0.8780 \\
       & R & 0.5897 & 0.7923 & 0.6000 & 0.7818 & 0.8889 & 0.7872 & 0.7399 \\
       & F1 & 0.7188 & 0.8747 & 0.6462 & 0.7350 & 0.9316 & 0.8810 & 0.7978 \\
\hline
ELECTRA & P & 0.9048 & 0.9911 & 0.8750 & 0.7183 & 0.9835 & 0.9348 & 0.9012 \\
        & R & 0.5135 & 0.8383 & 0.6562 & 0.9107 & 0.8775 & 0.9348 & 0.7885 \\
        & F1 & 0.6552 & 0.9084 & 0.7500 & 0.8031 & 0.9275 & 0.9348 & 0.8298 \\
\hline
T5 & P & 1.0000 & 1.0000 & 1.0000 & 1.0000 & 0.9545 & 1.0000 & 0.9924 \\
   & R & 0.4872 & 0.4214 & 0.5263 & 0.4590 & 0.0897 & 0.2034 & 0.3645 \\
   & F1 & 0.6552 & 0.5930 & 0.6897 & 0.6292 & 0.1641 & 0.3380 & 0.5115 \\
\hline
GPT-3 & P & 0.9333 & 0.9833 & 0.7143 & 0.6393 & 0.9595 & 1.0000 & 0.8716 \\
      & R & 0.3889 & 0.8217 & 0.5000 & 0.6724 & 0.8765 & 0.8298 & 0.6815 \\
      & F1 & 0.5490 & 0.8952 & 0.5882 & 0.6555 & 0.9161 & 0.9070 & 0.7518 \\
\hline
\end{tabular}
\end{table*}

\subsection{RELATIONSHIP BETWEEN ENTITY CATEGORIES AND THEIR AMOUNTS}
As shown in Figure 4, the performance of different models in identifying entities varies significantly, especially when compared to the frequency of occurrence for each entity type. Entities that occur more frequently, such as RRE (3674) and GPU (3606), generally have higher precision and recall scores across most models. This indicates that having a larger dataset contributes to better model performance. For example, models like BERT, RoBERTa, and ELECTRA show notably high F1 scores for entities like RRE and GPU. In a research paper, when examining various transformer models, including BERT, RoBERTa, and ELECTRA using a detailed emotion , it was found that the size of the model did not have a significant impact on the task of emotion recognition. This suggests that while data size may affect performance, the size and architecture of the models also play important roles \cite{bib65}.

However, there are some exceptions to this pattern. Despite having an equal number of occurrences in the data-set (570), both PNR and CMS exhibit fluctuating performance between categories. This variability suggests that the quantity alone is not the sole factor determining model performance; the complexity or uniqueness of the entity type may also play a role. For example, a comparative analysis also examined how well these models recognise emotions from texts, providing further insight into their performance on entity recognition tasks across different categories and data-sets. Additionally, a separate study focused on domain specific applications explored these models' ability to extract various clinical concepts, offering insights into their capacity to handle different types of entities and understanding how domain and data-set size can impact model performance \cite{bib66}.

RoBERTa consistently achieves the highest F1 score among the models, closely followed by BERT. This indicates that having a large amount of data can improve model performance, but the architecture and training techniques are still crucial factors. When it comes to tasks requiring a balanced precision and recall, RoBERTa or BERT are considered the most suitable options. Furthermore, a study on recognizing Protected Health Information (PHI) entities revealed differences in training times between these models, which could indirectly impact their performance on entity recognition tasks. These findings suggest that training time and computational resources may also influence how well different models perform in entity recognition tasks \cite{bib67}.

When using these models, it's important to consider both the frequency of the entity in the data and its complexity to ensure the best results. Research on NER using RoBERTa and ELECTRA models has shown that performance varies depending on the specific model and dataset used. For instance, an ELECTRA-based model performed better than BERT-based models when working with a dataset related to drugs, as measured by its F1 score. This highlights how the choice of model and characteristics of the dataset significantly impact entity recognition performance  \cite{bib68}. It underscores the importance of considering factors such as data availability, model architecture, and entity complexity in order to achieve optimal results in entity recognition tasks.
\begin{figure}[t!]
    \centering
    \includegraphics[width=1.2\textwidth]{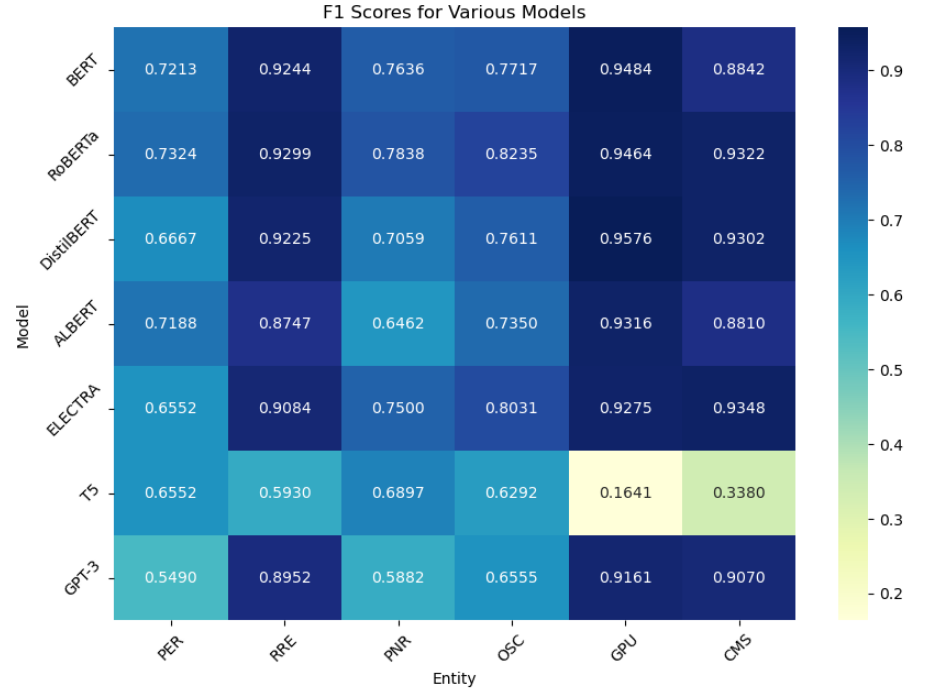} 
    \caption{\textbf{F1 scores of different models for each entity.}}
    \label{fig1}
\end{figure}

\subsection{IMPACT OF HYPER-PARAMETER FINE TUNING ON MODELS PERFORMANCE}
Grid search (GS) is a traditional technique used in fine-tuning parameters in machine learning and deep learning tasks, including NLP with popular models such as BERT, GPT, and T5. For instance, in a research, the authors used grid search to thoroughly refine BERT and other models using the DuoRC dataset. They focused on key hyper-parameters such as maximum sequence length, maximum question length, document stride, and training batch size, tweaking them prior to training to enhance model performance \cite{bib69}. Also, in common practice, GS is performed across a range of parameter values to figure out the combination that yields the best results for a given task. This approach plays an especially crucial role in the NLP field, where parameters like learning rate, batch size, and optimizer type can greatly affect the performance of models like BERT, GPT, and T5. The GS method works by thoroughly assessing models across a particular parameter grid, set up as follows: 
\begin{equation}
G = \{(p_1, p_2, \ldots, p_n) \mid p_i \in P_i\}
\tag{11}
\label{eq11}
\end{equation}
where \( P_i \) shows the set of possible values for hyper-parameter \( i \).

This study involves exploring different sets of hyper-parameters, including learning rates (\( lr \)), batch size (\( BS \)), epsilon (\( \varepsilon \)), and two unique optimizers—Adam and AdamW, as shown in Table 5. The set \( P_i \) represents the set of possible values for hyper-parameter \( i \). This section specifically looks at how these factors affect the overall performance of models in Named Entity Recognition (NER) tasks in construction supply chain risk management. This is distinct from the previous section, which evaluated the performance of models on individual entities.

\begin{table}[h]
\caption{\textbf{Models' Hyper-parameters}}\label{tab:hyperparameters}
\setlength{\tabcolsep}{3pt}
\begin{tabular}{@{}llp{6cm}@{}} 
\toprule
\textbf{Hyper-parameter} & \textbf{Values} & \textbf{Description} \\
\midrule
Learning Rate & 1e-6, 5e-6, 1e-5, 3e-5, 5e-5, 1e-4, 5e-4, 1e-3 & Step size for model weight updates. \\
Epsilon & 1e-7, 1e-8, 1e-9 & Small number to prevent any division by zero in the implementation. \\
Batch Sizes & 16, 32, 64 & Number of samples processed before the model is updated. \\
Optimizers & Adam, AdamW & Algorithms to change the attributes of the neural network such as weights and learning rate. \\
\botrule
\end{tabular}
\end{table}

Using GS, this study found 144 unique combinations to assess how different mixes of hyper-parameters affect model performance in NER tasks. Table VI shows the results of these assessments, pointing out the best combination for higher precision, recall, and F1 score, as well as the most efficient combination. It also considers the less successful combinations, giving a full view of performance across various hyper-parameter setups. When conducting hyper-parameter tuning, it's essential to focus on models that are most relevant and promising for the task at hand. In this context, we concentrated on transformer models like BERT, RoBERTa, DistilBERT, ALBERT, and ELECTRA, excluding T5 and GPT-3. This decision was based on several key considerations. First, transformer models have shown exceptional performance in understanding context and generating language, making them ideal for a wide range of NLP tasks. Each of these models, from BERT's pioneering architecture to ELECTRA's efficiency in understanding language, has unique strengths that make them suitable for in-depth hyper-parameter optimization. Second, due to their architecture and training methods, models like T5 and GPT-3 require significantly more computational resources for training and tuning, which may not be feasible or necessary for the specific objectives of our project. Moreover, GPT-3's closed-source nature and licensing limitations also posed a constraint. Therefore, our focus on the selected transformer models was driven by a balance of performance, resource availability, and specific model characteristics that align with our project goals. The best hyper-parameter combinations are shown in Table 6.

\begin{table*}[htbp]
\caption{Best Hyper-parameter Combinations Based on GS}
\label{table:hyperparameter_results}
\centering
\setlength{\tabcolsep}{3pt}
\begin{tabular}{@{}lllllllll@{}}
\toprule
\textbf{Models} & \multicolumn{4}{c}{\textbf{Hyper-parameters}} & \textbf{Precision} & \textbf{Recall} & \textbf{F1-score} & \textbf{Efficiency (s)} \\

 & \textbf{lr} & \textbf{BS} & \textbf{$\varepsilon$} & \textbf{Optimizer} & & & & \\
\midrule
BERT & 3e-05 & 16 & 1e-9 & Adam & 0.7882 & 0.8449 & 0.8097 & 161.2068 \\
 & 1e-4 & 32 & 1e-9 & Adam & 0.8032 & 0.7797 & 0.7824 & 145.9247 \\
 & 3e-05 & 16 & 1e-9 & Adam & 0.7882 & 0.8449 & 0.8097 & 161.2068 \\
 & 1e-3 & 64 & 1e-9 & Adam & 0.1372 & 0.1428 & 0.1399 & 134.9384 \\
\midrule
RoBERTa & 1e-3 & 64 & 1e-9 & Adam & 0.7753 & 0.8350 & 0.7944 & 106.7353 \\
 & 1e-3 & 64 & 1e-9 & Adam & 0.7753 & 0.8350 & 0.7944 & 106.7353 \\
 & 3e-05 & 32 & 1e-8 & AdamW & 0.7618 & 0.8412 & 0.7903 & 116.7825 \\
 & 1e-06 & 64 & 1e-8 & Adam & 0.6850 & 0.7692 & 0.7138 & 106.5951 \\
\midrule
DistilBERT & 1e-3 & 16 & 1e-7 & AdamW & 0.7898 & 0.7963 & 0.7841 & 84.8277 \\
 & 5e-05 & 32 & 1e-9 & AdamW & 0.7985 & 0.7787 & 0.7725 & 75.5385 \\
 & 1e-3 & 64 & 1e-8 & AdamW & 0.7351 & 0.7998 & 0.7569 & 69.8459 \\
 & 1e-3 & 64 & 1e-9 & Adam & 0.1378 & 0.1428 & 0.1403 & 68.8453 \\
\midrule
ALBERT & 3e-5 & 32 & 1e-8 & AdamW & 0.8100 & 0.8029 & 0.8018 & 155.8440 \\
 & 1e-3 & 16 & 1e-9 & Adam & 0.8235 & 0.7424 & 0.7738 & 163.4706 \\
 & 5e-5 & 32 & 1e-7 & AdamW & 0.7900 & 0.83027 & 0.7994 & 155.8783 \\
 & 1e-3 & 64 & 1e-9 & Adam & 0.1378 & 0.14285 & 0.14032 & 147.5978 \\
\midrule
ELECTRA & 1e-3 & 32 & 1e-9 & AdamW & 0.7910 & 0.8054 & 0.7933 & 149.6477 \\
 & 1e-3 & 32 & 1e-9 & AdamW & 0.7910 & 0.8054 & 0.7933 & 149.6477 \\
 & 5e-5 & 16 & 1e-8 & AdamW & 0.7480 & 0.8201 & 0.7766 & 165.5166 \\
 & 1e-3 & 64 & 1e-9 & Adam & 0.1378 & 0.1428 & 0.1403 & 137.6781 \\
\botrule
\end{tabular}
\end{table*}

In the conducted grid search for named entity recognition within the context of Australian construction supply chain risk management, distinct trends and implications have been revealed through the comparative analysis of models such as BERT, RoBERTa, DistilBERT, ALBERT, and ELECTRA, in relation to various hyper-parameters and optimizers. It has been observed that competitive performance is exhibited by both BERT and RoBERTa, with BERT slightly outperforming in terms of recall, indicative of its effectiveness in identifying relevant entities. Conversely, RoBERTa is distinguished by offering a more balanced trade-off between precision and recall, coupled with higher efficiency, positioning it as a time-efficient alternative \cite{bib70}.

DistilBERT, characterized by its lighter architecture, has been noted for its efficiency, achieving this without significant sacrifices in precision and recall, thereby emerging as a robust option under constraints of computational resources or time. In a different vein, ALBERT has been recognized for its precision, especially under specific hyper-parameter configurations, rendering it particularly suitable for tasks where precise identification is critical. ELECTRA, while not outshining in specific metrics, has been acknowledged for providing a consistent balance across various performance measures, which can be advantageous in scenarios demanding uniform performance.

Further insights have been gained into the effects of hyper-parameters and optimizers, where the learning rate has been identified as a critical factor influencing model performance. Generally, lower learning rates have been found to yield better recall and F1-scores, suggesting the benefit of a cautious approach in weight updating within this specific domain. However, it has been noted that excessively low learning rates might impair the learning capabilities of the model. Consistently, larger batch sizes have been associated with diminished performance, indicating the effectiveness of smaller batch sizes for this particular application.

Regarding the choice of optimizer, no consistent preference has been discerned between Adam and AdamW. However, it has been observed that models employing AdamW, particularly in the cases of DistilBERT and ALBERT, demonstrate enhanced efficiency. This improved efficiency might be attributable to the weight decay strategy of AdamW, which aids in expediting the convergence process. Hence, the findings underscore the necessity for a tailored selection of models and hyper-parameters in named entity recognition tasks, with the aim of aligning them with the specific requirements of the task at hand. BERT and RoBERTa have been noted for their proficiency in recall, while DistilBERT and ALBERT excel in efficiency and precision, respectively. ELECTRA, as a model, stands out for its well-rounded performance. The employment of lower learning rates and smaller batch sizes has generally been found to be more effective, while the choice between Adam and AdamW optimizers appears to be more influenced by considerations of efficiency than by factors of precision, recall, or F1-score. Figure 5 shows comparative analysis of transformer models’ performance using Adam and AdamW optimizers across various hyper-parameters.

When assessing precision, recall, and F1 scores in relation to optimizer choice, it is apparent that AdamW tends to enhance model performance across most models, suggesting its superiority in handling weight decay and perhaps aiding in generalization. However, the degree of this enhancement varies, indicating differing levels of sensitivity among the models to the optimization method. Considering learning rates, a lower learning rate coupled with AdamW optimizer seems to consistently benefit models like BERT, ALBERT, and to some extent, RoBERTa, in achieving higher F1 scores. DistilBERT and Electra, on the other hand, exhibit a less pronounced preference, indicating a potential robustness to a wider range of learning rates or an architecture that is less amenable to the subtle improvements offered by AdamW’s weight decay.

The impact of batch size is another critical factor. Larger batch sizes with AdamW appear to be particularly beneficial for models like BERT and ALBERT, as seen in their F1 scores. This could imply that these models, when paired with AdamW, are better able to capitalize on the stability provided by larger batches. Conversely, DistilBERT and Electra do not exhibit a strong preference, which might point to an intrinsic efficiency in these models that makes them less dependent on batch size for performance improvements. Epsilon values, which provide numerical stability in the optimization process, do not show a clear trend across the models, suggesting that its impact might be overshadowed by the more dominant effects of learning rates and batch sizes. In direct comparison, BERT and ALBERT show a strong preference for the AdamW optimizer, especially in larger batch sizes and lower learning rates, indicating their reliance on finer optimization techniques for peak performance. RoBERTa, while also benefiting from AdamW, does not display as stark a difference, which may imply an inherent robustness in the model’s architecture. DistilBERT’s and Electra’s performances, less affected by the optimizer choice, suggest that these models may have an intrinsic resilience to the optimization process, potentially due to their simpler or more efficient pre-training strategies.

In summary, while AdamW generally provides a performance edge, the extent of its benefits varies by model, with BERT and ALBERT showing the greatest improvements, RoBERTa demonstrating moderate sensitivity, and DistilBERT and Electra indicating a more optimizer-agnostic behavior. This reflects the complex interplay between model architecture and optimization techniques, underscoring the necessity for model-specific hyper-parameter tuning.

\begin{figure*}[htbp!]
\centering
\includegraphics[width=1.1\textwidth]{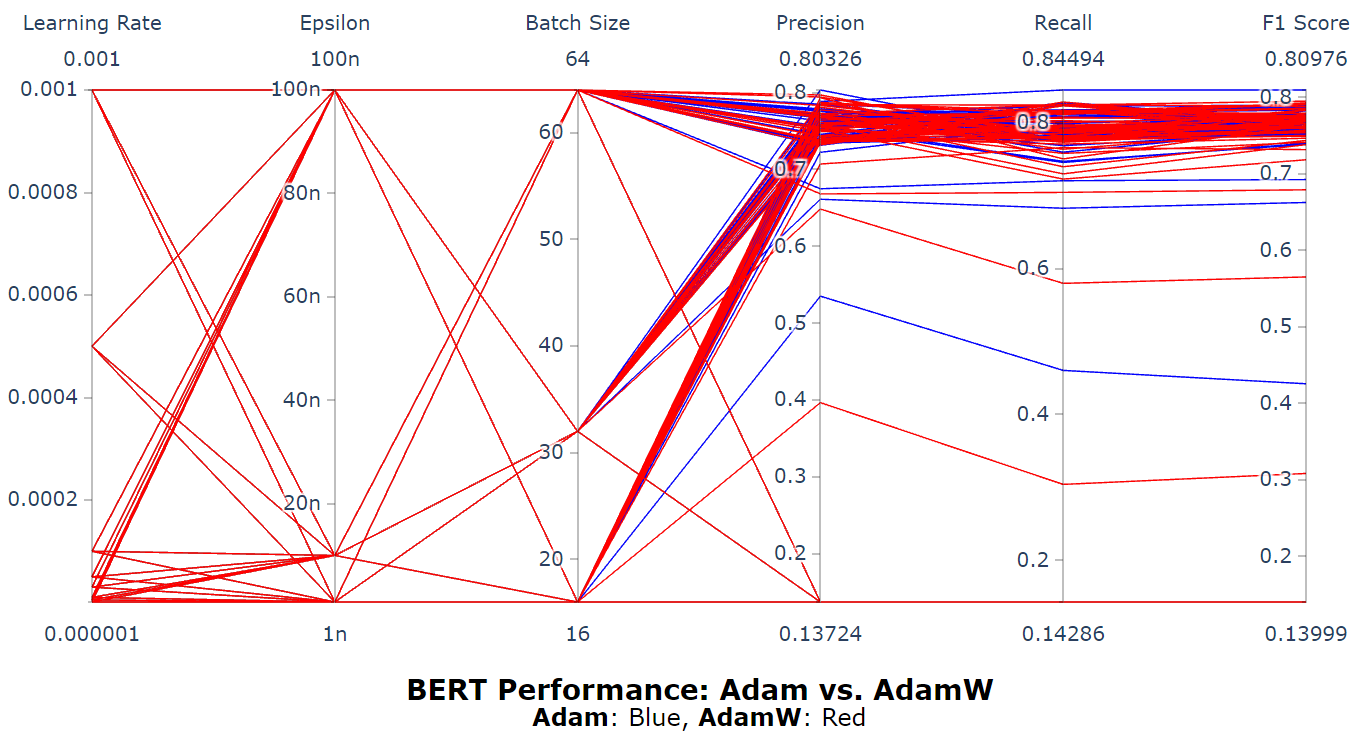}
\par\medskip
\includegraphics[width=1.1\textwidth]{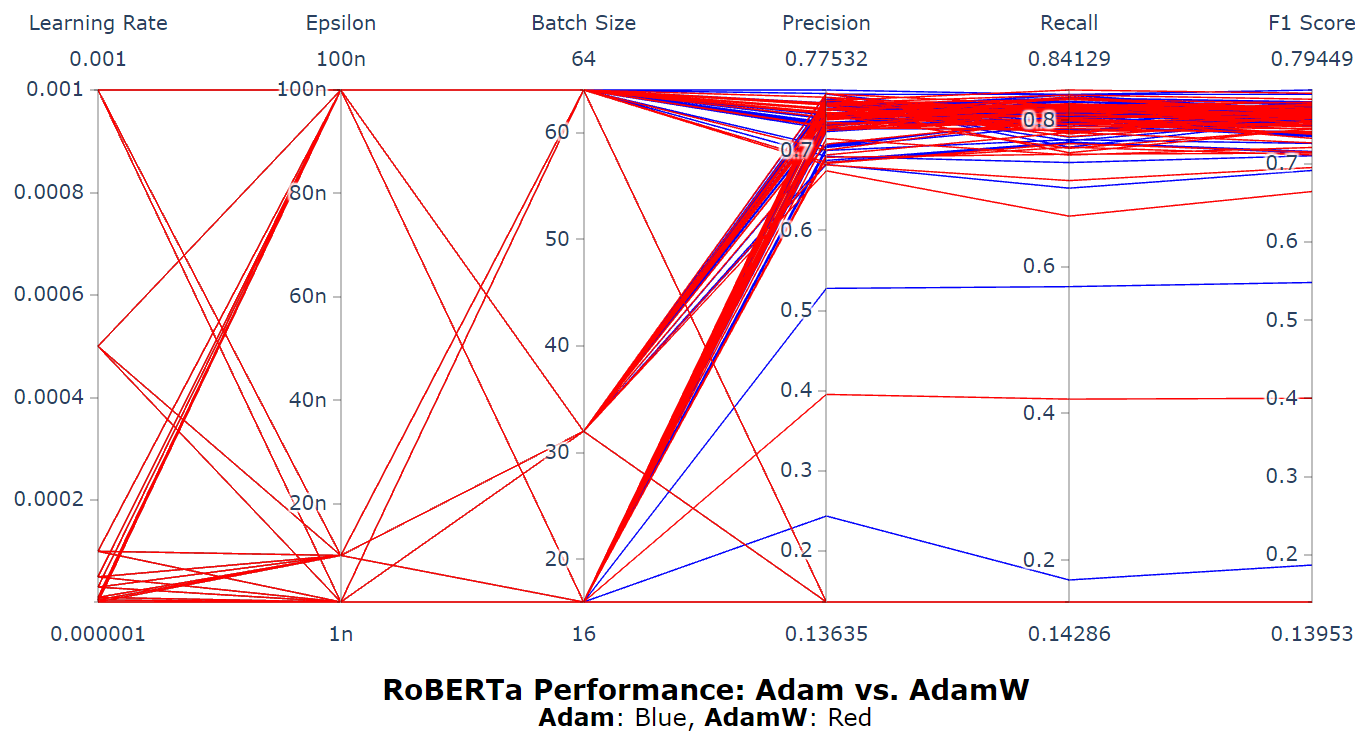}
\end{figure*}

\begin{figure*}[htbp!]
\centering
\includegraphics[width=1.1\textwidth]{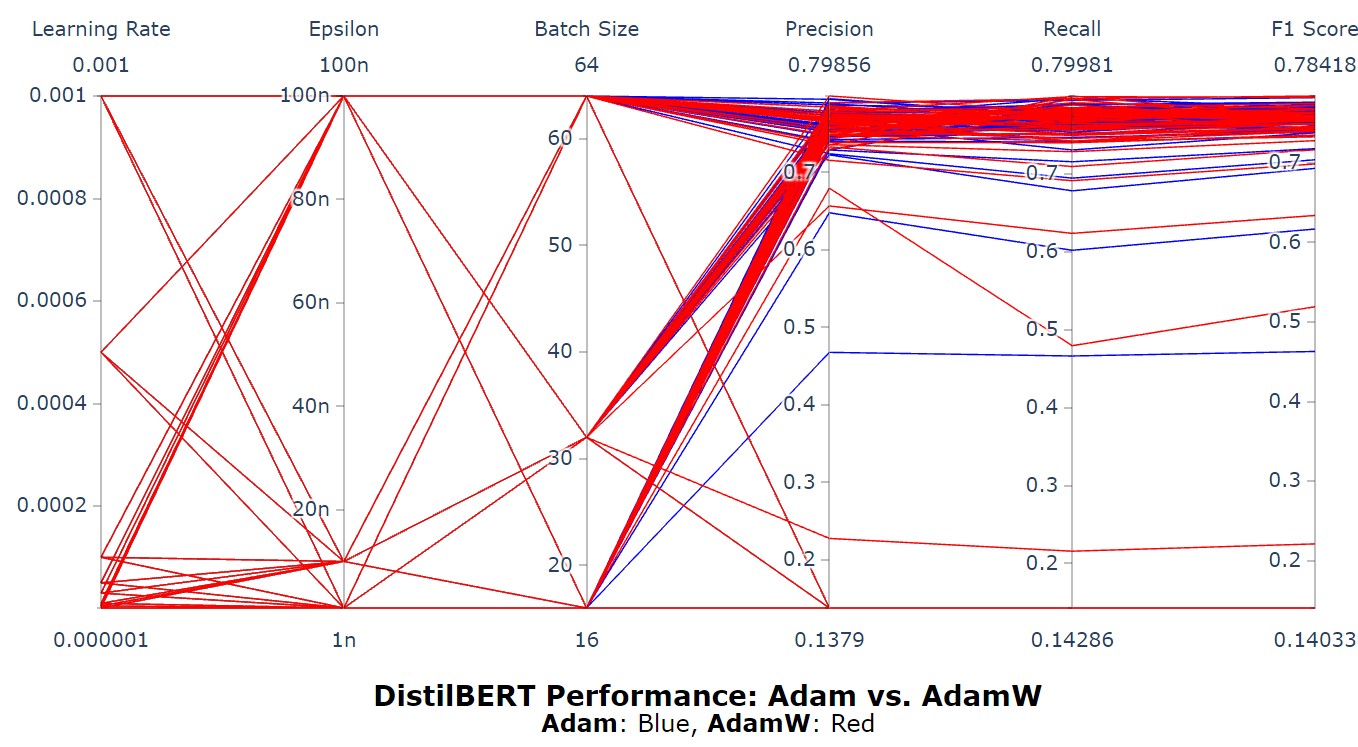}
\par\medskip
\includegraphics[width=1.1\textwidth]{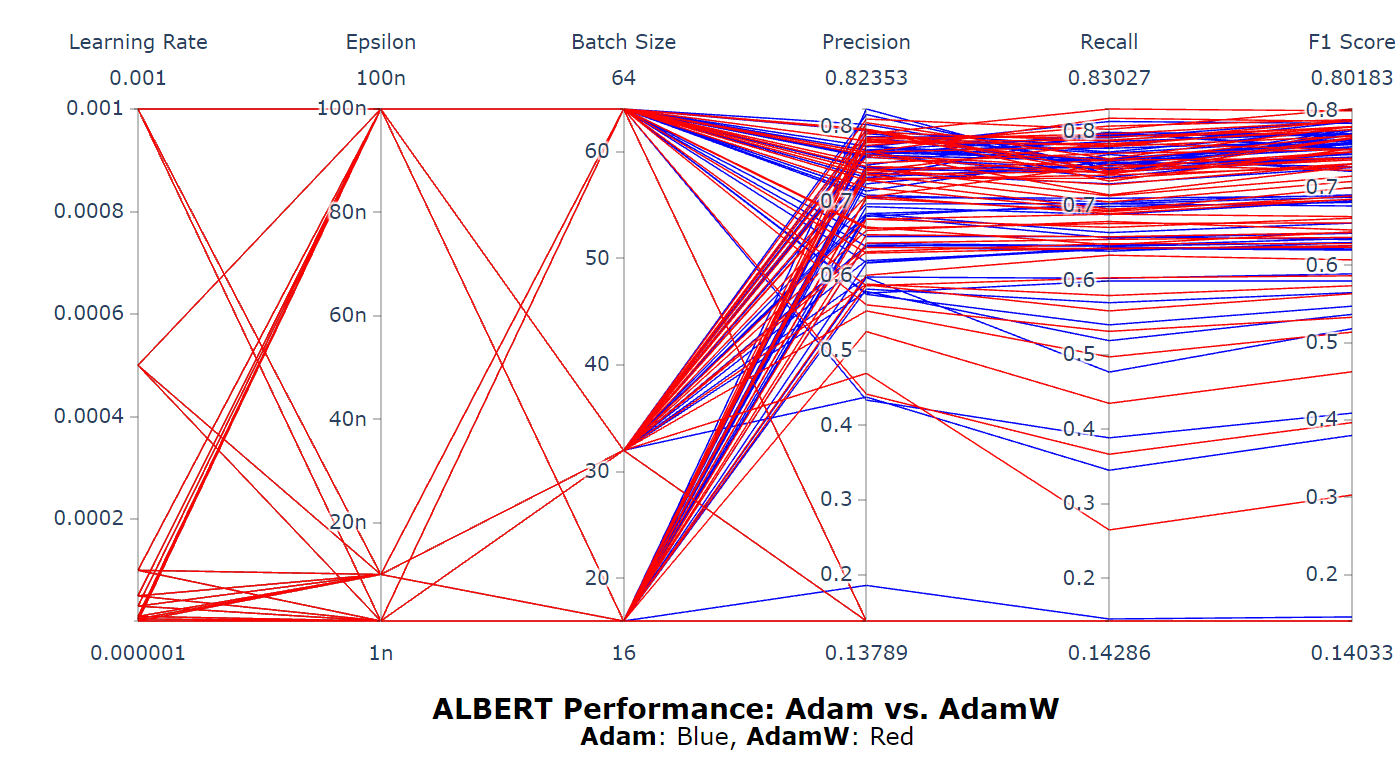}
\end{figure*}

\clearpage 

\begin{figure*}[htbp!]
\centering
\includegraphics[width=1.1\textwidth]{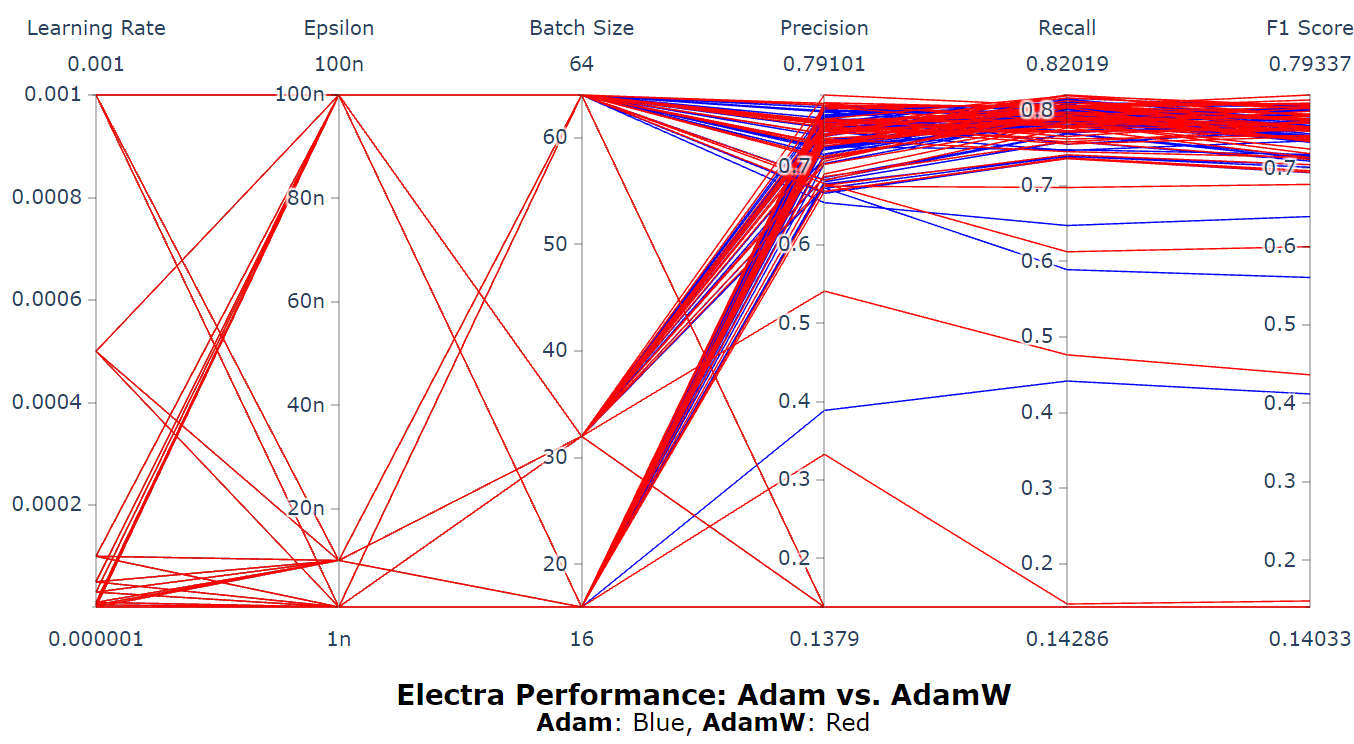}
\caption{\textbf{Comparative Analysis of Transformer Models’ Performance Using Adam and AdamW Optimizers Across Various Hyperparameters.}}
\label{fig:all_images}
\end{figure*}

\section{Conclusions and Future Directions}\label{sec5}
This study has demonstrated the effectiveness of various transformer-based models in Named Entity Recognition (NER) within the Australian construction supply chain risk management context, specifically using news articles. Models like BERT, RoBERTa, DistilBERT, ALBERT, and ELECTRA were evaluated, highlighting their respective strengths in processing and analyzing textual data for risk identification and management. The findings underscore the importance of NER in enhancing supply chain resilience and proactive risk management in the construction industry. A limitation of this study is the exclusion of the T5 and GPT-3 models from the grid search analysis.

\subsection{MODEL PERFORMANCE}
The comparative analysis of different transformer models revealed varying levels of efficacy in NER tasks. Models like BERT and RoBERTa showed robust performance, particularly in terms of precision and recall, indicating their suitability for extracting relevant entities from complex textual data. These insights are crucial for advancing the field of NER and its application in construction supply chain risk management.

\subsection{Project Management Performance}
Sophisticated transformer models like BERT, RoBERTa, and ELECTRA have revolutionised project management. They can analyze large amounts of text data, provide detailed risk profiles, and empower project managers to make proactive decisions. With these tools, project planning becomes more agile, allowing managers to navigate risk factors efficiently and stay on track with timelines and budgets. NER technologies have a transformative impact on project management in construction. By using NER to analyse global news trends, project managers can detect early warning signs of potential disruptions in the construction supply chain. This is especially important in the Australian construction sector, where external factors like international market dynamics and geopolitical shifts can have a significant impact. With NER, project managers can anticipate and plan for these risks, ensuring that projects stay on course even amidst global uncertainties. This approach increases resilience and enhances project execution.

\subsection{Future Directions}

In the exploration of alternative transformer models, further studies can consider trying out XLNet and DeBERTa. XLNet uses a unique permutation-based training method that could help us better understand the sequence of events in construction projects. On the other hand, DeBERTa has a disentangled attention mechanism that could enhance the recognition of entities from complicated construction documents.

•	 To further improve the performance of NER models in construction risk management, a more detailed strategy for tuning hyper-parameters could be employed. This involves expanding the scope of the grid search to consider a wider range of parameters for the Adam and AdamW optimizers. For example, different weight decay rates or learning rate schedules could be explored. 

•	The integration of NER capabilities into project management software could greatly improve the risk identification process. This integration would provide project managers with real-time alerts and suggestions, leveraging the latest news and market trends. As a result, project management becomes more reactive and adaptive.

•	Sentiment analysis is a valuable tool for risk assessment. By combining NER with sentiment analysis, we can gain a better understanding of the potential impact of identified risks. By assessing the sentiment of news articles and reports, we can prioritise risks based on their urgency.

•	 The creation of collaborative AI systems that can engage multiple stakeholders has the potential to democratise the risk assessment process. By incorporating input from various experts and enabling them to train and fine-tune the NER systems, future studies can develop a more robust model that is better suited to the specific requirements of different projects.


\bibliographystyle{plain}  
\bibliography{sn-bibliography}  

\end{document}